\documentclass[amsmath, amssymb, aps, pra,reprint, twocolumn,longbibliography,floatfix,nofootinbib,superscriptaddress]{revtex4}
\usepackage[utf8]{inputenc}
\usepackage[urlcolor=blue, colorlinks=true,citecolor=blue]{hyperref}

\usepackage{amsmath}
\usepackage{amsmath,amssymb}
\usepackage{graphicx}
\usepackage{textcomp, gensymb}
\usepackage{tabularx}
\usepackage{algorithm}
\usepackage{algpseudocode}
\usepackage{bm}

\newcommand{\ket}[1]{\left|#1\right\rangle}
\newcommand{\bra}[1]{\left\langle #1\right|}

\begin{document}
\title{Hybrid quantum physics-informed neural networks\\ for simulating computational fluid dynamics in complex shapes}

\author{Alexandr~Sedykh}
\address{Terra Quantum AG, 9000 St.~Gallen, Switzerland}

\author{Maninadh~Podapaka}
\address{Evonik Operations GmbH, 63450 Hanau-Wolfgang, Germany}

\author{Asel~Sagingalieva}
\address{Terra Quantum AG, 9000 St.~Gallen, Switzerland}

\author{Karan~Pinto}
\address{Terra Quantum AG, 9000 St.~Gallen, Switzerland}

\author{Markus~Pflitsch}
\address{Terra Quantum AG, 9000 St.~Gallen, Switzerland}

\author{Alexey~Melnikov}
\thanks{Corresponding author, e-mail: alexey@melnikov.info
\begin{center}
\fbox{
\begin{minipage}{0.45\textwidth}
Please check the published version, which includes all the latest additions and corrections: Mach. Learn.: Sci. Technol. 5:025045, 2024, DOI: \href{https://doi.org/10.1088/2632-2153/ad43b2}{10.1088/2632-2153/ad43b2}
\end{minipage}
}
\end{center}
}
\address{Terra Quantum AG, 9000 St.~Gallen, Switzerland}


\begin{abstract}
Finding the distribution of the velocities and pressures of a fluid by solving the Navier-Stokes equations is a principal task in the chemical, energy, and pharmaceutical industries, as well as in mechanical engineering and the design of pipeline systems. With existing solvers, such as OpenFOAM and Ansys, simulations of fluid dynamics in intricate geometries are computationally expensive and require re-simulation whenever the geometric parameters or the initial and boundary conditions are altered. Physics-informed neural networks are a promising tool for simulating fluid flows in complex geometries, as they can adapt to changes in the geometry and mesh definitions, allowing for generalization across fluid parameters and transfer learning across different shapes. We present a hybrid quantum physics-informed neural network that simulates laminar fluid flows in 3D $Y$-shaped mixers. Our approach combines the expressive power of a quantum model with the flexibility of a physics-informed neural network, resulting in a 21\% higher accuracy compared to a purely classical neural network. Our findings highlight the potential of machine learning approaches, and in particular hybrid quantum physics-informed neural network, for complex shape optimization tasks in computational fluid dynamics. By improving the accuracy of fluid simulations in complex geometries, our research using hybrid quantum models contributes to the development of more efficient and reliable fluid dynamics solvers.
\end{abstract}

\maketitle

\section{Introduction}
Computational fluid dynamics (CFD) solvers are primarily used to find the distribution of the velocity vector, $\bm{v}$, and pressure, $p$, of a fluid (or several fluids) given the initial conditions (e.g., initial velocity profile) and the geometrical domain in which the fluid flows~\cite{cfd_review, Anderson1995ComputationalFD}. 
To do this, it is necessary to solve a system of differential equations~\cite{Simmons1972DifferentialEW} called the Navier-Stokes (NS) equations that govern the fluid flow~\cite{Jameson2009MeshlessMF}. A well-established approach is to use numerical CFD solvers from several vendors, as well as publicly accessible alternatives, such as OpenFOAM~\cite{openfoam} or Ansys~\cite{ansys}. These solvers discretize a given fluid volume into several small parts known as cells~\cite{Unstructured}, where it is easier to get an approximate solution and then join the solutions of all the cells to get a complete distribution of the pressure and velocity over the entire geometrical domain. 

While this is a rather crude explanation of how CFD solvers work, discretizing a large domain into smaller pieces accurately captures one of their main working principles~\cite{Mari2013voFoamA}.
The runtime of the computation and the accuracy of the solution both sensitively depend on the fineness of discretization, with finer grids taking longer but giving more accurate solutions.
Furthermore, any changes to the geometrical parameters necessitate the creation of a new mesh and a new simulation. This process consumes both time and resources since one has to remesh and rerun the simulation every time a geometrical parameter is altered~\cite{pinn_ns_review}.

We propose a workflow employing physics-informed neural networks (PINNs)~\cite{raissi2019physics} to escape the need to restart the simulations whenever a geometrical property is changed completely. A PINN is a new promising tool for solving all kinds of parameterized partial differential equations (PDEs)~\cite{pinn_ns_review}, as it does not require many prior assumptions, linearization or local time-stepping. One defines an architecture of the neural network (number of neurons, layers, etc.) and then embeds physical laws and boundary conditions into it via constructing an appropriate loss function, so the prior task immediately travels to the domain of optimization problems.

For the classical solver, in the case of a parameterized geometric domain problem, getting accurate predictions for new modified shapes requires a complete program restart, even if the geometry has changed slightly. In the case of a PINN, to overcome this difficulty, one can use the transfer learning method~\cite{Transfer_Learning} (Sec.~\ref{sec:transfer}), which allows a model previously trained on some geometry to be trained on a slightly modified geometry without the need for a complete reset.

Also, using a trained PINN, it is easy to obtain a solution for other parameters of the PDE equation (e.g. kinematic viscosity in the NS equation~\cite{Jameson2009MeshlessMF}, thermal conductivity in the heat equation, etc.) with no additional training or restart of the neural network, but in the case of traditional solvers, restarts cannot be avoided.

One of the features that makes PINNs appealing is that they suffer less from the curse of dimensionality. Finite discretization of a $d$-dimensional cube with $N$ points along each axis would require $N^d$ points for a traditional solver. In other words, the complexity of the problem grows exponentially as the sampling size $d$ increases. Using a neural network, however, one can define a $\mathbb{R}^d \rightarrow \mathbb{R}$ mapping (in case of just one target feature) with some weight parameters. Research on the topic suggests that the amount of weights/complexity of a problem in such neural networks grows polynomially with the input dimension $d$~\cite{hutzenthaler2020proof, grohs2018proof}. This theoretical foundation alone allows PINNs to be a competitive alternative to solvers.

It is worth noting that although a PINN does not require $N^d$ points for inference, it does require many points for training. There exist a variety of sampling methods, such as latin hypercube sampling~\cite{stein1987large}, Sobol sequences~\cite{sobol2011construction}, etc., which can be used to improve a PINN's convergence on training~\cite{zubov2021neuralpde}. However, a simple static grid of points is used in this work for the purpose of simplicity.

Classical machine learning can benefit substantially from quantum technologies. In \cite{Gaitan2020}, quantum computing is used in a similar problem setting. The performance of the current classical models is constrained by the high computing requirements. Quantum computing models can improve the learning process of existing classical models~\cite{dunjko2018machine, qml_review_2023, Neven2012QBoostLS, PhysRevLett.113.130503, saggio2021experimental, kordzanganeh2023parallel, kurkin2023forecasting}, allowing for better target function prediction accuracy with fewer iterations~\cite{asel1}. In many industries, including the pharmaceutical~\cite{sag_hyb_2022, fedorov}, aerospace~\cite{rainjonneau2023quantum}, automotive~\cite{asel2}, logistics~\cite{haboury2023supervised}and financial~\cite{Alcazar.Perdomo-Ortiz.2020,Coyle.Kashefi.2021,Pistoia.Yalovetzky.2021wtn,Emmanoulopoulos.Dimoska.2022n6o,Cherrat.Pistoia.2023} sector quantum technologies can provide unique advantages over classical computing. Many traditionally important machine learning domains are also getting potential benefits from utilizing quantum technologies, e.g., in image processing~\cite{senokosov2024quantum,Li.Wang.2022,naumov2023tetra,Riaz.Hopkins.2023} and natural language processing~\cite{Hong.Xiao.2022,Lorenz.Coecke.20218h,Coecke.Toumi.2020,Meichanetzidis.Coecke.202020f}. Solving nonlinear differential equations is also an application area for quantum algorithms that use differentiable quantum circuits~\cite{quantum_expressivity, paine2023physicsinformed} and quantum kernels~\cite{PhysRevA.107.032428}.

Recent developments in automatic differentiation enable us to compute the exact derivatives of any order of a PINN, so there is no need to use finite differences or any other approximate differentiation techniques. It, therefore, seems that we do not require a discretized mesh over the computational domain.
However, we still need a collection of points from the problem's domain to train and evaluate a PINN.
For a PINN to provide an accurate solution for a fluid dynamics problem, it is important to have a high expressivity (ability to learn solutions for a large variety of, possibly complex, problems). 
Fortunately, expressivity is a known strength of quantum computers~\cite{mo_2022, schuld2021effect, schuld2020circuit}. Furthermore, quantum circuits are differentiable, meaning their derivatives can be calculated analytically, which is essential for noisy intermediate-scale quantum devices.

\begin{figure*}[t!]
    \centering
    \includegraphics[width=1\linewidth]{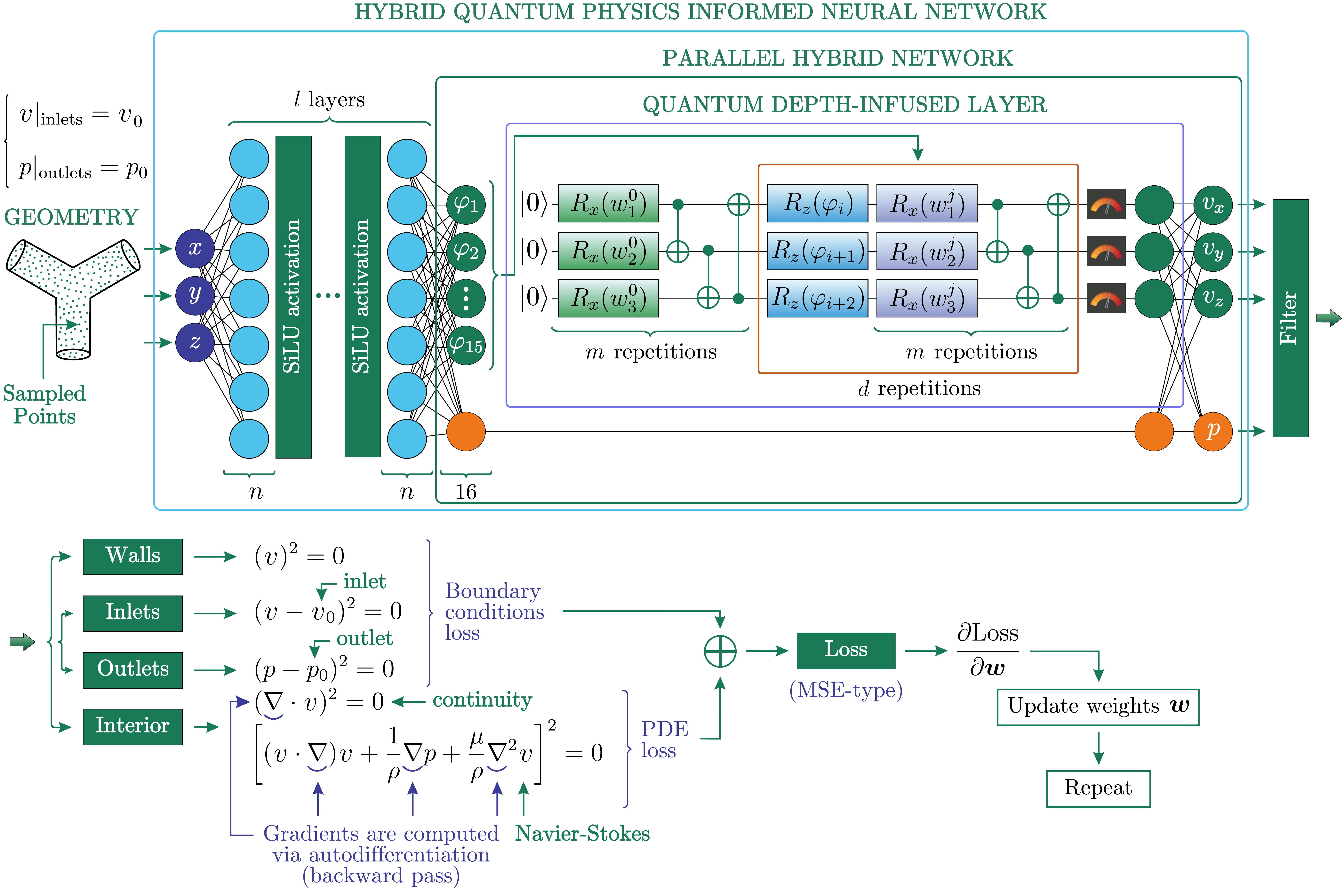}
    \caption{ 
    Scheme for the implementation of the HQPINN's training process and its architecture. HQPINN takes the $(x, y, z)$ coordinates of the points, sampled in the geometrical domain (purple circles), and yields $(\bm v, p)$ as output, where $\bm v$ is the velocity vector (green circles) and $p$ is the pressure (orange circle). The neural network itself begins with a classical part, which is a multilayer perceptron with $l$ layers of $n$ neurons.
    Then, there is a Parallel Hybrid Network, which consists of a Quantum depth-infused layer (a variational quantum circuit) and a multilayer perceptron.  A ``Filter'' mechanism divides input points into four groups—fluid domain, walls, inlets, and outlets—each with its specific constraint (a boundary condition or a PDE). The error for each constraint is meticulously calculated and incorporated into the total loss, which is then minimized via gradient descent to refine the model's predictions. The process iteratively adjusts the network's weights using full-batch gradient descent, ensuring that every point in the geometrical domain contributes to a holistic solution that respects both the fluid dynamics within the domain and the conditions at its boundaries.  The training process is described in Sec.~\ref{sec:classic_details}.} 
    \label{fig:hybrid_arc}
\end{figure*}

In this article, we propose a hybrid quantum PINN (HQPINN) shown in Fig.~\ref{fig:hybrid_arc} to solve the NS equations with a steady flow in a 3D $Y$-shape mixer. The general principles and loss function of PINN workflow are described in Sec.~\ref{sec:pinns}. The problem description, including the geometrical details, is presented in Sec.~\ref{sec:problem} while in Sec.~\ref{sec:classic_pinn} and Sec.~\ref{sec:hybrid_pinn}, we describe classical and hybrid PINNs in detail. Sec.~\ref{sec:classic_details} explains the intricacies of PINN's training process and simulation results. A transfer learning approach, applied to PINNs, is presented in Sec.~\ref{sec:transfer}. 
Conclusions and further plans are described in Sec.~\ref{sec:discussion}. 

\section{Physics-informed neural networks for solving partial differential equations}\label{sec:pinns}

PINNs were originally introduced in~\cite{raissi2019physics}. The main idea is to use a neural network - usually a feedforward neural network like a multilayer perceptron - as a trial function for a PDE's solution. 
Let us consider an abstract PDE:
\begin{equation}\label{eq:abs_pde}
    \mathcal{D}[f(\bm{r}, t); \lambda] = 0,
\end{equation}
where $\Omega \subset \mathbb{R}^d$ is the computational domain, $\bm{r} \in \Omega$ is a coordinate vector, $t \in \mathbb{R}$ is time, $\mathcal{D}$ is a nonlinear differential operator with $\lambda$ standing in for the physical parameters of the fluid and $f(\bm{r}, t)$ is a solution function.

Let us consider a neural network $u(\bm{r}, t)$ that takes coordinates, $\bm{r}$, and time, $t$, as input and yields some real value (e.g., the pressure of a liquid at this coordinate at this particular moment).

We can evaluate $u(\bm{r}, t)$ at any point in the computational domain via a forward pass and compute its derivatives (of any order) $\partial_t^n u(\bm r, t)$, $\partial_{\bm r}^n u(\bm r, t)$ through backpropagation~\cite{rumelhart1986learning}. Therefore, we could substitute $f(\bm{r}, t) = u(\bm{r}, t)$ and try to learn the correct solution for the PDE via common machine learning gradient optimization methods~\cite{gradient} (e.g., gradient descent).

This approach is inspired firstly by the ability to calculate the \textit{exact} derivatives of a neural network via autodifferentiation~\cite{autodiffPaper} and secondly by neural networks being universal function approximators~\cite{Hornik1989MultilayerFN}. 

The loss, $\mathcal{L}$, that the PINN tries to minimize is defined as
\begin{equation}
    \mathcal{L} = \mathcal{L}_\text{PDE} + \mathcal{L}_\text{BC},
\end{equation}
where $\mathcal{L}_\text{BC}$ is the boundary condition loss and $\mathcal{L}_\text{PDE}$ is the partial differential equation loss.

The boundary condition loss is responsible for satisfying the boundary conditions of the problem (e.g., a fixed pressure on the outlet of a pipe). For any field value, $u$, let us consider a Dirichlet (fixed-type) boundary condition~\cite{greenshields2022notes}
\begin{equation}
    u(\bm{r}, t)|_{\bm{r} \in B} = u_0(\bm{r}, t),
\end{equation}
where $u_0(\bm r, t)$ is a boundary condition and $B \subset \mathbb{R}^d$ is the region where a boundary condition is applied.

If $u(\bm{r}, t)$ is a neural network function (see Sec.~\ref{sec:pinns}), the boundary condition loss is calculated in a mean-squared error (MSE) manner: 
\begin{equation}\label{eq:bc_loss}
    \mathcal{L}_\text{BC} = \langle (u(\bm{r}, t) - u_0(\bm{r}, t))^2 \rangle_B,
\end{equation}
where $\langle \cdot \rangle_B$ denotes averaging over all the data points $\bm{r} \in B$ that have this boundary condition.

The PDE loss is responsible for solving the governing PDE. If we have an abstract PDE (\ref{eq:abs_pde}) and a neural network function $u(\bm{r}, t)$, substituting $f(\bm{r}, t) = u(\bm{r}, t)$ and calculating the mean-squared error of the PDE gives:
\begin{equation}\label{eq:pde_loss}
    \mathcal{L}_\text{PDE} = \langle (\mathcal{D}[u(\bm{r}, t); \lambda])^2 \rangle_\Omega,
\end{equation}
where $\langle \cdot \rangle_\Omega$ means averaging over all the data points in the domain of the PDE.

\section{Simulations}\label{sec:problem}

In this work, we consider the \textit{steady} (i.e., time-independent) flow of an \textit{incompressible} fluid in 3D without any external forces.

The NS equations (\ref{eq:ns}) and the continuity equation (\ref{eq:continuity}) describe this scenario as follows:
\begin{equation}\label{eq:ns}
    -(\bm{v} \cdot \nabla) \bm{v} + \nu \Delta \bm{v} - \frac{1}{\rho} \nabla p = 0,
\end{equation}
\begin{equation}\label{eq:continuity}
    \nabla \cdot \bm{v} = 0,
\end{equation}
where $\bm{v}(\bm{r})$ is the velocity vector, $p(\bm{r})$ is the pressure, $\nu$ is the kinematic viscosity and $\rho$ is the fluid density. The PDE parameters $\nu$ and $\rho$ were previously referred to as $\lambda$.
For each of the 4 PDEs (3 projections of vector equation \ref{eq:ns} and 1 scalar equation \ref{eq:continuity}), the $\mathcal{L}_\text{PDE}$ is calculated separately and then summed up.

\subsubsection{Training physics-informed neural networks}\label{sec:classic_details}

The geometry is a collection of points organized in a \textit{.csv} file. It is split into four groups: fluid domain, walls, inlets, and outlets.
The fluid domain is the domain in which the NS equations are solved, i.e., where the fluid flows. The other three groups have boundary conditions described in Sec.~\ref{sec:pinns}.

While untrained, PINN produces some random distribution of velocities and pressures. These values and their gradients are substituted into the corresponding NS equations and boundary conditions.
With every iteration, the weights of the neural network are updated to minimize the error in the governing equations, and our solution becomes more and more accurate.

The training iteration is simple: the point cloud is passed through the PINN, the MSE loss is calculated (getting it requires taking gradients of ($\bm{v}, p$) at each point of the geometry), the gradient of the loss with respect to the weights is taken and the parameters are updated.

To visualize the training outcomes of the neural network, we used ParaView~\cite{paraview}, and the simulation results are shown in Fig.~\ref{fig:30_35} (1) for loss and velocity distribution.

\subsubsection{Cylinder flow simulation}\label{sec:cylinder_flow}

At first, we used a simple 3D cylinder flow as a baseline solution for classical PINN. We also validated PINN's predictions by comparing them to OpenFOAM's solution.

\begin{figure}[h!]
    \centering
    \includegraphics[width=0.5\linewidth]{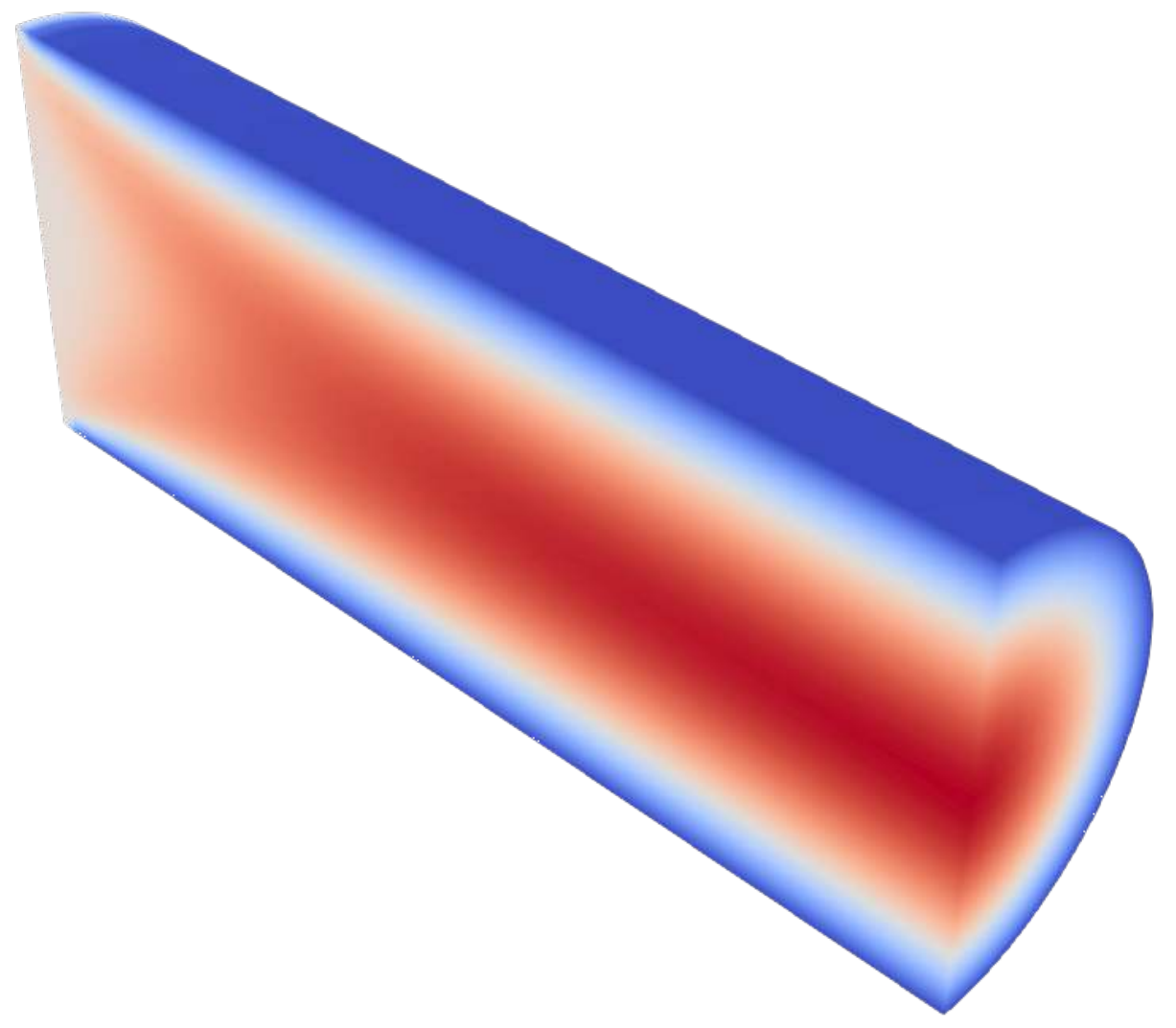}
    \caption{
    Isometric view of a slice of the cylinder. Color denotes the velocity magnitude for clarity.
    }
    \label{fig:3d_example}
\end{figure}

In this simulation, we have no-slip boundary condition on the walls, fixed velocity of $v_0 = 10 \ \text{mm}/\text{s}$ on the inlet, and fixed zero pressure $p = 0 \ \text{Pa}$ on the outlet.
The fluid is water with standard density $\rho = 1 \ \text{g} / \text{cm}^3$ and $\nu = 1 \ \text{mm}^2 / \text{s}$ kinematic viscosity. 
The cylinder has radius of $2 \ \text{mm}$ and height of $10 \ \text{mm}$ (\ref{fig:3d_example}). Reynolds number for such flow is $10$, so it can be considered laminar.

The training was done on a single NVIDIA A100 GPU for $20'000$ iterations with an L-BFGS optimizer. For its training, the model uses $25'000$ points inside the cylinder, which are taken straight from the mesh nodes used in OpenFOAM's simulation.
PINN represented the solver's solution quite well: the relative error of velocity magnitude averaged over the whole cylinder is $1.2 \ \%$, and the mean relative error of pressure is $0.8 \ \%$. Refer to Fig.~\ref{fig:cylinder} for ground truth (OpenFOAM) and prediction (PINN) field values, as well as the distribution of relative error across the geometry.

The only error spikes both for pressure and velocity fields are located near the inlet edge of the pipe, where the uniform velocity profile of $10 \ \text{mm}/\text{s}$ abruptly changes to $0$ due to non-slip boundary condition on walls, which is somewhat unphysical. This can be corrected by making the inlet velocity profile parabolic instead of uniform, but we decided to stick to a simple problem statement for the baseline solution.

\subsubsection{Generalization of physics-informed neural networks}

To check if our baseline model possesses any generalization capabilities, we trained the same model as in Section~\ref{sec:cylinder_flow}, but now incorporating $4$ input parameters: the $x, y, z$ coordinates and the kinematic viscosity $\nu$, which was held constant at $1\, \text{mm}^2/\text{s}$ for the aforementioned model. Given that we employ a full-batch strategy, where the optimizer processes the gradient of all points in the geometry simultaneously, such a modification significantly increases the memory requirements for storing the total gradient. Consequently, we limited our training set to a set of $\nu$ values, $\{1, 2, 3, 4\}\, \text{mm}^2/\text{s}$. Nonetheless, we believe this range is sufficient to explore the generalization capabilities of the classical PINN.

Following the same training regime as in Section~\ref{sec:cylinder_flow}, we present the results on the test data in Table~\ref{table:generalization}. The results indicate that the model generalizes well across the range of $\nu$ values it trained on.

\begin{table}[h!]
    \begin{tabular}{|l|l|l|}
    \hline
    $\nu, \ \text{mm}^2/\text{s}$ & $p$ rel. error, \% & $\|\bm{v}\|$ rel. error, \% \\ \hline
    $1  $   & $10.0$                    & $4.7$ \\ \hline
    $\bm{1.5}$   & $2.0 $                    & $2.8$ \\ \hline
    $2  $   & $2.7 $                  & $1.5$ \\ \hline
    $\bm{2.5}$   & $5.9 $                  & $3.6$ \\ \hline
    $3  $   & $8.5 $                  & $5.8$ \\ \hline
    $\bm{3.5}$   & $10.5$                  & $7.7$ \\ \hline
    $4  $   & $12.3$                  & $9.4$ \\ \hline
    $\bm{5}  $   & $15.1$                  & $12.3$ \\ \hline
    $\bm{10} $   & $21.0$                    & $24.6$ \\ \hline
    \end{tabular}
    \caption{
        Pressure $p$ and velocity magnitude $\|\bm{v}\|$ relative errors (averaged over the whole geometrical domain) between PINN predicted solution and reference OpenFOAM solution for different values of kinematic viscosity $\nu$. The values, which model did not train on, are put in bold.
    }
    \label{table:generalization}
\end{table}

\begin{figure*}[t]
    \centering
    \includegraphics[width=1\linewidth]{Fig_3.pdf}
    \caption{
    (a), (c), (e) Ground truth pressure and velocity projections. (b), (d), (f) PINN predicted pressure and velocity projections. (g), (h) Relative pressure and velocity magnitude errors between PINN and ground truth (absolute error divided by maximum pressure or velocity magnitude across the cylinder).
    }
    \label{fig:cylinder}
\end{figure*}

However, when the model attempts to extrapolate solutions for $\nu$ values outside its training range, the quality of the solutions begins to degrade rapidly. More specifically, the solution tends to converge to the trivial case of $\bm{v} = 0$ as it approaches the end of the pipe. This phenomenon will be revisited and discussed further in Section~\ref{sec:y-shape}, where we attempt simulations on more complex geometries than the current one.

\subsubsection{Y-shaped mixer flow simulation}\label{sec:y-shape}

\begin{figure*}[t!]
    \centering
    \includegraphics[width=1\linewidth]{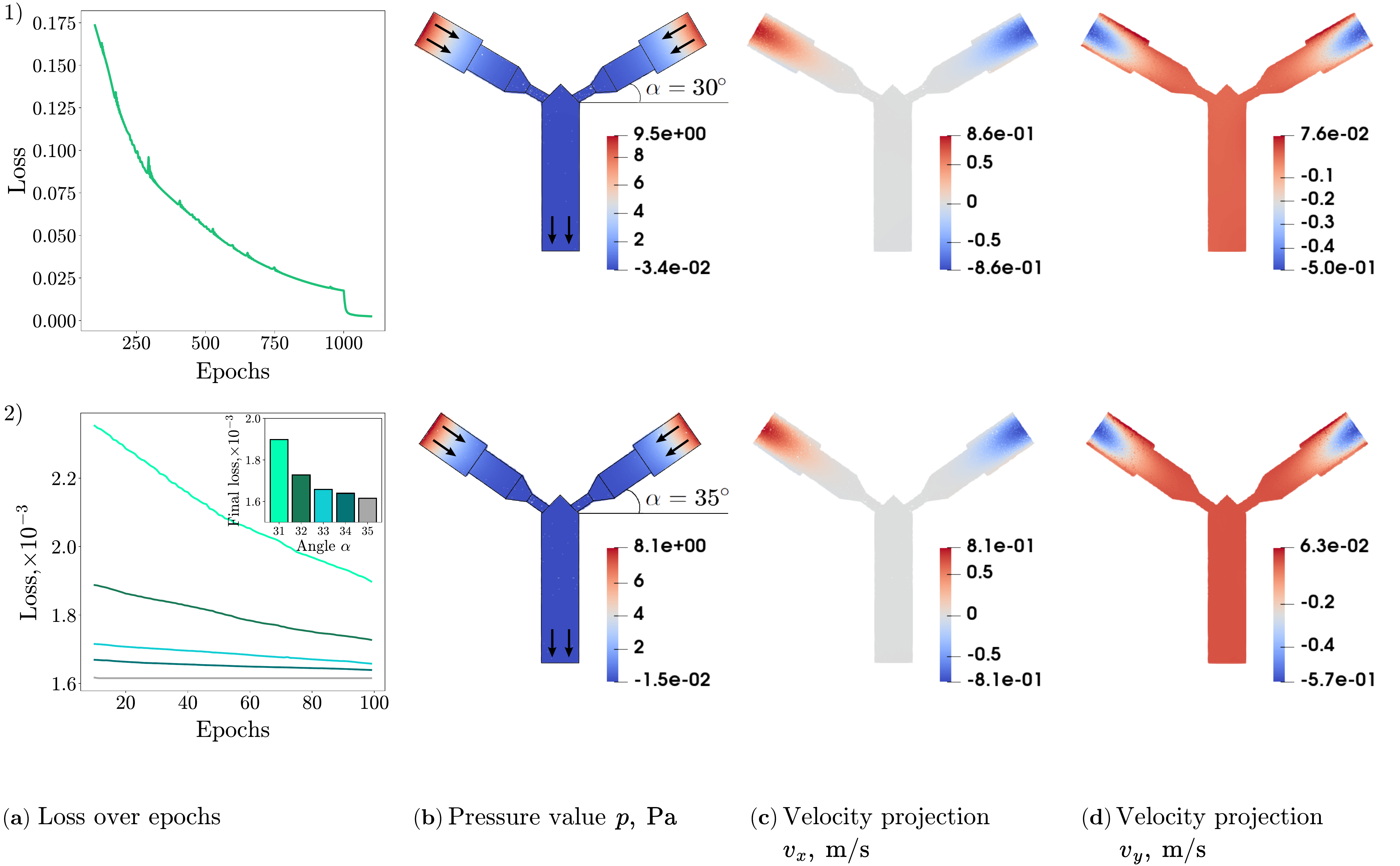}
    \caption{1) Classical model. (a) Loss curve. The classical PINN managed to learn a solution near the entry point well. However, it vanishes to zero further down the mixer.
    2) Transfer learned models. (a) Loss curve. The transfer-learned models trained better with each iteration, surpassing the original model. (b, c, d) Distributions of the fluid pressure and velocities for the last model with $\alpha = 35^\circ$.
    }
    \label{fig:30_35}
\end{figure*}

This time, we try to simulate the flow of liquid in a $Y$-shaped mixer consisting of three tubes (Fig.~\ref{fig:30_35}). The mixing fluids are identical and have parameters $\rho = 1.0 \ \text{kg}/\text{m}^3$ and $\nu = 1.0 \ \text{m}^2/\text{s}$.

The imposed boundary conditions are as follows: 
\begin{itemize}
    \item no-slip BC on walls: $\bm{v}(\bm{r})|_{\text{walls}} = 0$,
    \item fixed velocity profile on inlets: $\bm{v}(\bm{r})|_\text{inlets} = \bm{v}_0(\bm{r})$, 
    \item fixed pressure on outlets: $p(\bm{r})|_\text{outlets} = p_0$,
\end{itemize}
where, $\bm{v}_0(\bm{r})$ is a parabolic velocity profile on each inlet and $p_0(\bm{r}) = 0$.

The PINN was trained via full-batch gradient descent with the Adam optimizer~\cite{adam} for $1000$ epochs and then with the L-BFGS optimizer for $100$ epochs until gradient vanished and training became impossible. After the training, the PINN managed to learn a non-trivial downward flow at the beginning of both pipes. On the edges of these pipes, the velocities become zero, as they should, due to no-slip wall boundary conditions. However, further down the mixer, the solution degenerates to zero, so it does not even reach the mixing point. This fact should at least violate the continuity equation because there is an inward flow without matching outward flow. This problem is primarily caused by gradient vanishing inherent to PINN models. 

In an attempt to overcome the challenge, we introduce an HQPINN model in \ref{sec:hybrid_pinn}, which shows better results in terms of PDE and boundary conditions satisfaction. We also research potential ways of making a generalizable PINN in \ref{sec:transfer} by employing a transfer learning method. 

\subsubsection{Physics-informed neural network architecture}\label{sec:classic_pinn}

In this section, we provide details on the PINN's architecture.
The core of the PINN is a neural network whose architecture is a multilayer perceptron with several fully connected layers. As shown in Fig.~\ref{fig:hybrid_arc}, the first layer consists of $3$ neurons (since the problem is 3D), then there are $l = 5$ hidden layers with $n$ neurons, where $n = 128$ for the cylinder flow and $n = 64$ for the $Y$-shaped mixer. For the classical PINN, the ``Parallel Hybrid Network'' box is replaced with one fully connected layer $n \rightarrow 4$. There is no quantum layer in the classical case, so the output goes straight into the filter.
Between adjacent layers, there is a sigmoid linear unit (SiLU) activation function~\cite{elfwing2018sigmoid}. The PINN takes the $(x, y, z)$ coordinates as inputs and yields $(\bm{v}, p)$ as its output, where $\bm{v}$ is the velocity vector with three components and $p$ is the pressure.

\subsection{Transfer Learning}\label{sec:transfer}

Transfer learning is a powerful method of using the knowledge and experience of one model that has been pretrained on one problem to solve another problem~\cite{Transfer_Learning, mari2020transfer}. It is extremely useful because it means that a second model does not have to be trained from scratch. This is especially helpful for fluid modeling, where selecting the most appropriate geometrical hyperparameters would otherwise lead to the simulations being rerun many times.

For transfer learning, we used a model from the previous section as a base, which had $\alpha_0 = 30^{\circ}$, where $\alpha$ is the angle between the right pipe and the $x$ axis (see Fig.~\ref{fig:30_35}). Then, for each $\alpha = \{31^{\circ}, 32^{\circ}, 33^{\circ}, 34^{\circ}, 35^{\circ}\}$, we tried to obtain the solution, each time using the previously trained model as an initializer. For example, to transfer learn from $31^\circ$ to $32^\circ$, we used the $31^\circ$ model as a base and so on. Each iteration is trained for $100$ epochs with L-BFGS.
Fig.~\ref{fig:30_35} shows that the PINN adapts well to changes in the value of $\alpha$. That is, our hypothesis was correct: with PINNs, one does not need to completely rerun the simulation on a parameter change, transfer learning from the base model will suffice.

\subsection{Hybrid quantum PINN}\label{sec:hybrid_pinn}

\begin{figure*}[ht!]
    \centering
    \includegraphics[width=1\linewidth]{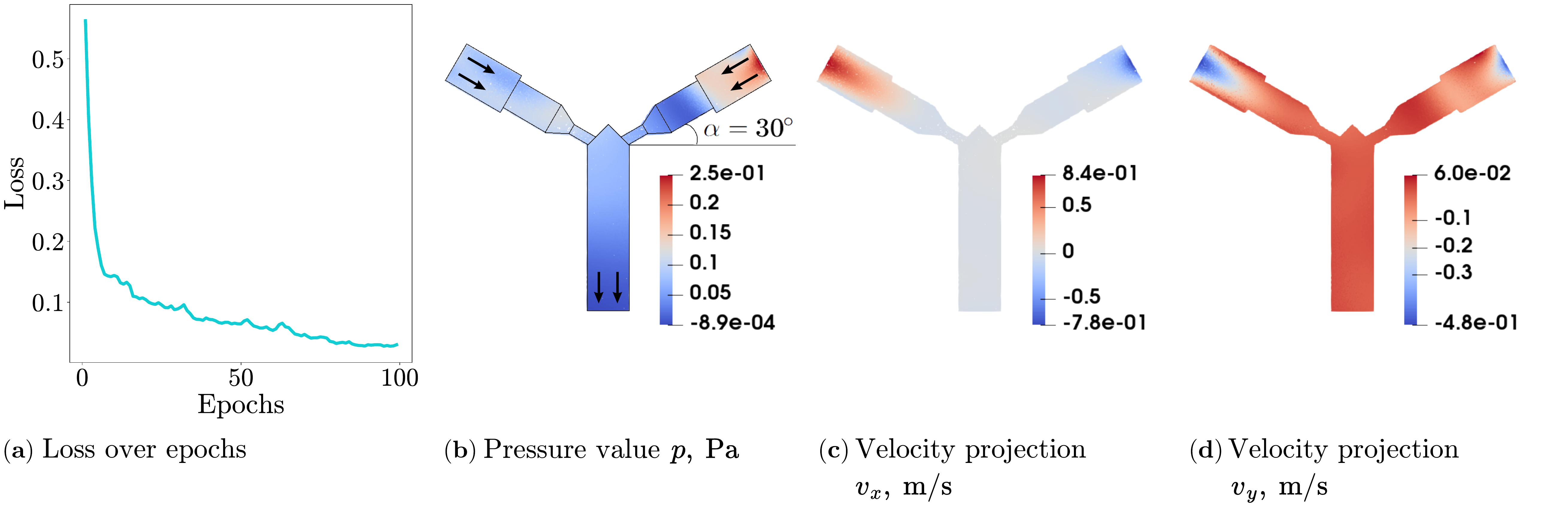}
    \caption{Hybrid quantum model. (a) Loss curve. (b, c, d) Distribution of the fluid pressure and velocities for the HQPINN at ($\alpha = 30^\circ$). Similarly to the classical model, a non-trivial solution near the entry point is present. However, there is an asymmetry between the left and the right pipe, which should not be there. This could be an effect of the data encoding strategy.
    }
    \label{fig:30_q}
\end{figure*}

Quantum machine learning was shown to be applicable to solving differential equations~\cite{quantum_expressivity, paine2023physicsinformed, PhysRevA.107.032428}. Here, we introduce a hybrid quantum neural network called HQPINN and compare it against its classical counterpart, classical PINN. As shown in Fig.~\ref{fig:hybrid_arc}, the architecture of HQPINN is comprised of classical fully connected layers, specifically a multilayer perceptron, coupled with a Parallel Hybrid Network~\cite{kordzanganeh2023parallel}. The latter is a unique intertwining of a Quantum depth-infused layer~\cite{sag_hyb_2022, schuld2021effect} and a classical layer. Interestingly, the first 15 units ($\phi_1, \dots , \phi_{15}$, depicted in green) of the layer are dedicated to the quantum layer, whereas the information from one orange unit proceeds along the classical pathway. This parallel structure enables simultaneous information processing, thereby enhancing the efficiency of the learning process.

Regarding the quantum depth-infused layer, it is implemented as a variational quantum circuit (VQC), a promising strategy for navigating the complexities of the Noisy Intermediate-Scale Quantum (NISQ) epoch~\cite{preskill2018quantum}. The NISQ era is characterized by quantum devices with a limited number of qubits that cannot yet achieve error correction, making strategies like VQCs particularly relevant~\cite{zhao2019qdnn, dou2021unsupervised, sag_hyb_2022}.

The capacity of HQPINN to direct diverse segments of the input vector toward either the quantum or classical part of the network equips PINNs with an enhanced ability to process and learn from various patterns more efficiently. For instance, some patterns may be optimally processed by the quantum network, while others might be better suited for the classical network. This flexibility in processing contributes to the robust learning capabilities of HQPINN.

\subsubsection{Quantum Depth-Infused Layer}\label{sec:vqc}

Transitioning into the building blocks of our model, the quantum depth-infused layer takes center stage. Quantum gates are the basic building blocks for any quantum circuit, including those used for machine learning. Quantum gates come in single-qubit (e.g., rotation gate $R_y(\theta)$, gate that plays a central role in quantum machine learning) and multiple-qubit gates (e.g., CNOT) and modulate the state of qubits to perform computations. The $R_y(\theta)$ gate rotates the qubit around the y-axis of the Bloch sphere by an angle $\theta$, while the two-qubit CNOT gate changes the state of one qubit based on the current state of another qubit. Gates can be fixed, which means they perform fixed calculations, such as the Hadamard gate, or they can be variable, such as the rotation gate that depends on the rotation angle and may perform computations with tunable parameters.

To extract the results, qubits are measured and projected onto a specific basis, and the expected value is calculated. When using the $\sigma_z$ Pauli matrix observable, the expected value of the measurement is represented as: $\bra{\psi} \sigma_z \ket{\psi}$, with $\psi$ signifying the wave function that depicts the current state of our quantum system. For a more detailed understanding of quantum circuits, including logic gates and measurements, standard quantum computing textbooks such as~\cite{nielsen2002quantum} offer a comprehensive guide.

To refine the functioning of the network we propose, the initial phase of data processing is performed on a classical computer. Subsequently, these data are integrated into the quantum gate parameters in an encoding layer (blue gates) repeated $d=5$ times, forming part of the quantum depth-infused layer followed by the variational layer (green gates). Number of repetitions of the variational layer $m = 2$. As the algorithm develops, these gate parameters in the variational layer are dynamically adjusted. Finally, at the measurement level, the qubits are quantified, leading to a series of classical bits as the output.

Conceptually, a quantum algorithm resembles a black box that receives classical information as input and emits classical information as output. The objective here is to fine-tune the variational parameters such that the measurement outcome most accurately reflects the prediction function. In essence, this parameter optimization is akin to optimizing the weights in a classical neural network, thereby effectively training the quantum depth-infused layer.

\subsubsection{Training hybrid quantum physics-informed neural network}\label{sec:training_quantum}

The HQPINN consists of the classical PINN with weights pre-initialized from the previous stage (as described in Sec.~\ref{sec:classic_pinn}), a parallel hybrid network and a fully-connected layer at the end. 

The training process of the hybrid PINN does not differ from that of the classical PINN except in the following ways. Firstly, all calculations are done on the classical simulator of quantum hardware, the QMware server~\cite{qmw_qmw_2022}, which has recently been shown to be quite good for running hybrid algorithms~\cite{mo_bench_2022}.

Secondly, how does one backpropagate through the quantum circuit layer? The answer is to use the ``adjoint differentiation'' method, introduced in~\cite{adjointdiff}, which helps to compute derivatives of a VQC on a classical simulator efficiently.

\begin{figure}[!ht]
    \centering
    \includegraphics[width=0.97\linewidth]
    {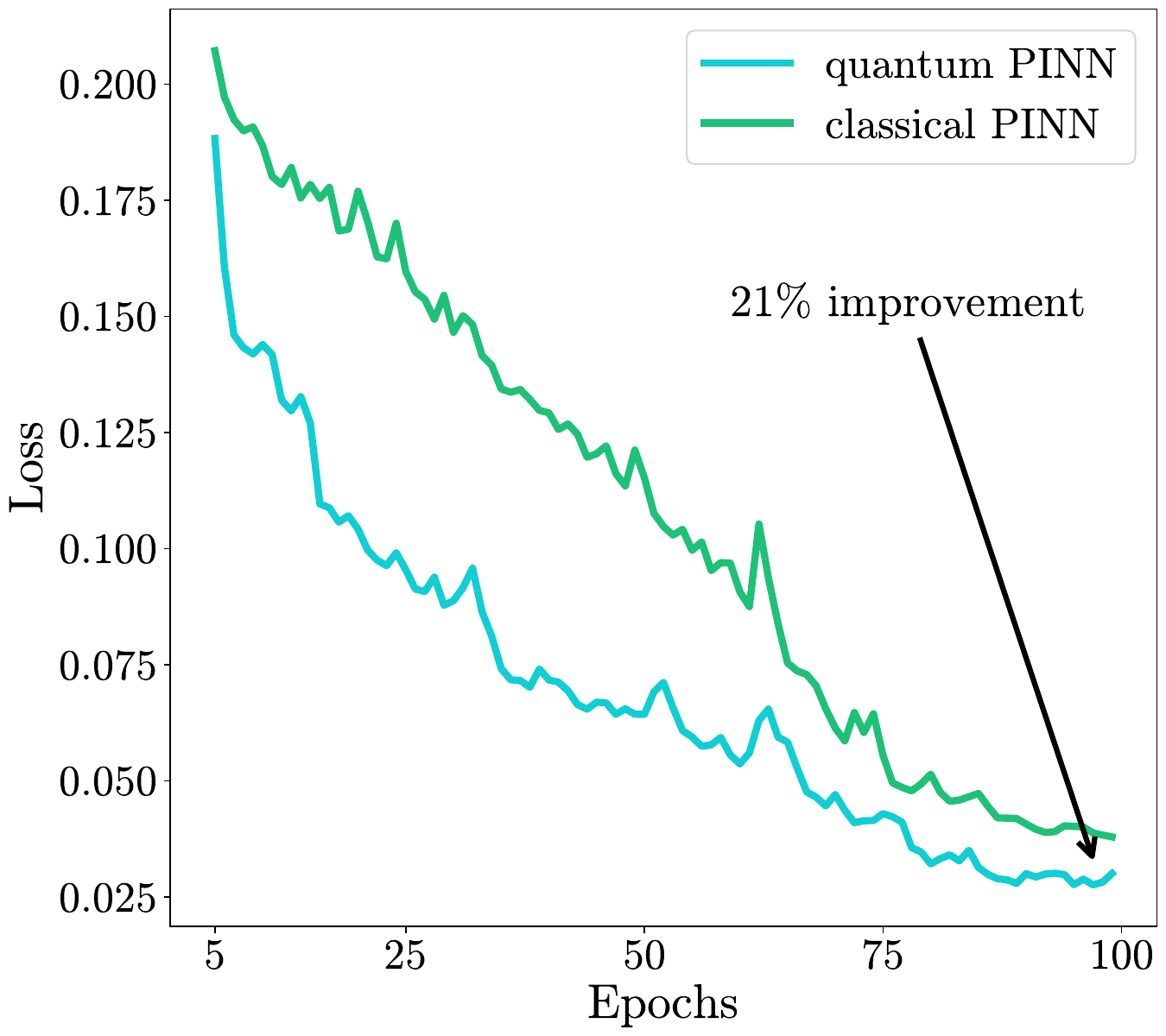}
    \caption{
    Losses for the classical and hybrid quantum PINNs.
    HQPINN outperforms the classical PINN in terms of loss value due to higher expressivity.
    }
    \label{fig:loss_qc_comparison}
\end{figure}

This time the model was trained for $100$ epochs using mini-batch gradient descent with the Adam optimizer (Fig.~\ref{fig:30_q}). The mini-batch strategy was employed due to the very low training speed of quantum circuits, as they train on a CPU. We will then compare this model with a purely classical one, with the same architecture from Sec.~\ref{sec:classic_pinn}, but this time trained only with mini-batch Adam. All learning hyperparameters (learning rate, scheduler parameters, batch size) are shared between the quantum and classical models. Comparing the two, Fig.~\ref{fig:loss_qc_comparison} shows that the quantum model outperforms the classical one in terms of the loss value by $21 \%$. As the loss function for PINNs directly corresponds to PDE and BC satisfaction, it implies that HQPINN has achieved better physical accuracy before the gradient vanish. However, the mini-batch training strategy for the HQPINN is far from ideal and stems from the large training time of simulated quantum circuits. Proper hardware backend for highly parallel GPU-accelerated quantum computing could greatly extend the HQPINN advantage.

Additionally, the question of computational demands scaling is of importance. For the CFD examples considered in the paper, we used a three-qubit circuit, which is not computationally demanding to execute on a classical computer. However, with each additional qubit, the runtime of the quantum circuit execution will approximately double. That means, according to the benchmarks~\cite{mo_bench_2022}, the runtime will become significant beyond 20 qubits circuits. Nonetheless, with the development of quantum computers, the scaling is expected to be more favorable as the runtime on a quantum chip would stay constant in case the circuit depth is fixed. Therefore, we believe that for more complex shapes, the simulation would require more data and, hence, larger HQPINN models require the need of quantum computers for simulation.

\section{Discussion}\label{sec:discussion}

In our investigation, we pursued two distinct goals aimed at advancing the capabilities of PINNs through the integration of quantum computing methodologies. The first objective focused on enhancing the performance of classical PINNs in solving PDEs within 3D geometries through the adoption of HQPINNs. This endeavor was motivated by the recognized success of classical PINNs in 2D problem settings~\cite{raissi2019physics} and the existing gap in their application to more complex 3D scenarios. The significance of this goal lies in its potential to broaden the applicability of PINNs to a wider range of scientific and engineering problems characterized by three-dimensional spaces.

Our second goal was to investigate the presence and feasibility of transfer learning in classical PINNs when applied to 3D geometries. This exploration is crucial for understanding the ability of PINNs to generalize across different geometrical configurations and physical problems, potentially enabling more efficient re-training processes on novel problems and geometries. The overarching aim here is to enhance the model's versatility and reduce computational costs associated with the training of neural networks for new tasks.

Our results in applying HQPINNs to 3D PDE problems demonstrate a notable improvement in reducing PDE loss, marking a significant step forward in our first objective. Specifically, we observed a $21\%$ reduction in loss when comparing the performance of HQPINNs against their purely classical counterparts. This quantitative improvement highlights the enhanced expressiveness and computational efficiency brought about by the integration of quantum layers, suggesting that quantum computing elements can indeed augment the capability of PINNs in handling the intricacies of 3D problems. However, it is important to recognize that despite these advancements, achieving an optimal solution for 3D PDEs remains a challenging endeavor. The observed loss reduction, while substantial, does not resolve the complexities of 3D geometrical problem solving, indicating the need for further refinement and exploration of the HQPINN framework.

In the realm of transfer learning, our exploration yielded encouraging signs that classical PINNs possess an inherent capability to adapt to variations in a geometrical shape, when theangle of a $Y$-shaped mixer was changed. This qualitative finding is instrumental in demonstrating the potential for PINNs to be applied in shape optimization tasks and other applications requiring flexibility across different geometrical configurations. However, the journey towards realizing full transfer learning and generalization capabilities in PINNs is still in its early stages. Our initial successes serve as a foundation for further research, underscoring the necessity for continued development and investigation into the mechanisms that enable effective transfer learning within PINNs.

In developing our HQPINNs, we have also considered recent advances in other quantum and quantum-inspired methods for solving nonlinear equations, such as based on the matrix product states approach~\cite{gourianov2022quantum} and variational quantum algorithms for nonlinear problems~\cite{lubasch2020variational,quantum_expressivity}. While our approach primarily focuses on handling complex boundary conditions, the integration of these quantum-inspired methods and variational quantum algorithms could further improve the performance and accuracy of our proposed quantum PINN model.

Collectively, our findings contribute valuable insights into the potential of hybrid quantum neural networks and the prospects for transfer learning in solving complex physical problems. As we continue to push the boundaries of what is achievable with PINNs, our future efforts will focus on refining these models to approach accurate full-scale CFD simulation in complex 3D shapes.

The plan includes exploring better architectures for the quantum PINN, investigating their impact on expressiveness, generalizability and optimization landscape, and trying data-driven approaches. Entirely different networks, such as neural operators~\cite{neural_ops,fno} and graph neural networks~\cite{mesh_based_gnn, sanchez2020learning}, could also be considered in a quantum setting and enhanced with quantum circuits. 

\bibliography{lib}
\bibliographystyle{unsrt}

\end{document}